\documentclass[letterpaper, 10 pt, conference]{ieeeconf}  

\IEEEoverridecommandlockouts                              

\usepackage{graphics} 
\graphicspath{{figures/}} 
\usepackage{mathtools}
\usepackage{amsmath} 
\usepackage{amssymb}  
\usepackage{siunitx}
\let\labelindent\relax
\usepackage[inline]{enumitem}
\usepackage{booktabs}
\usepackage{threeparttable}

\newcommand{\norm}[1]{\left\lVert#1\right\rVert}
\DeclareMathOperator*{\argmin}{arg\,min} 

\title{\LARGE \bf
Active Constraints for Simultaneous Guidance and Safety in Handheld Micromanipulation
}

\author{Roger Ho, Robert A. MacLachlan, and Cameron N. Riviere
\thanks{This work was supported in part by the U.S. National Institutes of Health (grant number R01GM153059).}
\thanks{Department of Mechanical Engineering and Robotics Institute, CMU, Pittsburgh, PA 15213 USA (email: {\tt\small criviere@andrew.cmu.edu})}%
\thanks{This work has been submitted to the IEEE for possible publication. Copyright may be transferred without notice, after which this version may no longer be accessible.}
}

\begin{document}

\maketitle
\thispagestyle{empty}
\pagestyle{empty}

\begin{abstract}
In micromanipulation, erroneous hand motion limits unaided manipulation accuracy, which can affect task performance and safety in critical applications like microsurgery. Handheld robots (micromanipulators) enable motion accuracy beyond human limits by providing robotic assistance in a low-cost, compact, intuitive form factor. Particularly, active constraints are an effective form of robotic assistance, which aids the operator by restricting the robot's permissible motion based on task-specific requirements or objectives, thereby providing task guidance and enforcing safety. However, existing formulations of active constraints for handheld systems are limited, most notably because of the inability to provide guidance and enforce safety simultaneously. Implementing active constraints with optimization-based methods can address these limitations, but existing implementations are designed for free-standing robots, and neglect that handheld robots actuate relative to the operator's hand with a limited workspace. In this work, we propose an optimization-based active constraint framework for handheld micromanipulation that enables simultaneous guidance and safety enforcement. The formulation was implemented in a handheld robot and demonstrated on a medically relevant artificial task. Results show that the proposed formulation reduced safety violation rate by more than two orders of magnitude relative to baselines, while recovering the threefold accuracy loss from enforcing safety alone.
\end{abstract}

\section{Introduction}
\label{sec:intro}

In micromanipulation, unintentional, erroneous hand motion (physiological tremor, drift, jerk, etc.) limits unaided manipulation accuracy~\cite{singhPhysiologicalTremorAmplitude2002}. This positioning error of the hand not only degrades the efficacy of micromanipulation, but also concerns safety in critical applications like microsurgery. To enable accuracy beyond human limits, robotic assistance can be applied to micromanipulation to improve precision, accuracy, and reliability~\cite{taylorMedicalRoboticsComputerintegrated2003,roizenblattRobotassistedVitreoretinalSurgery2018}. Particularly, \textit{active constraints} (virtual fixtures) are an effective method of robotic assistance, which aid the operator by shaping the robot's permissible motion~\cite{bowyerActiveConstraintsVirtual2014a}. Given a task with safety requirements, active constraints can provide task guidance (e.g., by constraining the robot to move along a desired trajectory or guidance region) and/or enforce safety (e.g., by preventing the robot from entering unsafe/forbidden regions). 

Active constraints and other types of robotic assistance in micromanipulation are typically realized via three main modalities: teleoperation, where the operator controls a slave robot via a physically separate master robot; comanipulation, where the operator and the robot share control of the same physical end effector; and handheld, where the operator holds the entire robot, which actuates relative to the handle~\cite{maclachlanMicronActivelyStabilized2012a}. Of these modalities, handheld robots are attractive due to their familiar and intuitive interface at a lower cost and smaller footprint when compared to teleoperated and comanipulated robots, which are generally large, free-standing systems. However, advantages of the handheld form factor also introduce fundamental tradeoffs that affect how active constraints are implemented. 

Notably, handheld robots lack a mechanical ground separate from the operator, unlike teleoperated and comanipulated systems (hereafter collectively referred to as \textit{grounded systems}). Without a ground to react against, conventional force-based enforcement of active constraints~\cite{menothmohanReviewRecentMethods2025} cannot be implemented; handheld robots must enforce constraints positionally~\cite{beckerHandheldMicromanipulationVisionbased2011} within a workspace much smaller than that of grounded systems~\cite{maclachlanMicronActivelyStabilized2012a}. Because of these differences, formulations of active constraints for grounded systems cannot be generalized to handheld systems~\cite{beckerVisionBasedControlHandheld2013}.

At the same time, the active constraint frameworks developed specifically for handheld systems are limited. Most importantly, they can only enforce a single constraint at any given moment~\cite{beckerVisionBasedControlHandheld2013}, so task guidance cannot be provided simultaneously while enforcing safety. To address this gap, we propose an improved active constraint framework for handheld micromanipulation using constrained optimization. The contributions of this work are summarized as follows:
\begin{enumerate}
  \item We present an active constraint framework for handheld robots that can provide task guidance and enforce safety simultaneously, while degrading guidance gradually when actuation budget can no longer support both.
  \item We validate the formulation on a physical handheld system with a medically relevant artificial task, demonstrating a reduction in safety violation rate by more than two orders of magnitude compared to baselines, without the 3$\times$ accuracy loss incurred by enforcing safety alone.
\end{enumerate}

\section{Related Work}

Active constraints are conventionally enforced by shaping the forces between the operator and the robot~\cite{menothmohanReviewRecentMethods2025}. Optimization-based formulations compute the permitted motion by constrained optimization, which can enforce multiple constraints of different geometries~\cite{mingliConstrainedOptimizationApproach2005} and deal with hard limits such as joint and actuation bounds~\cite{marinhoUnifiedFrameworkTeleoperation2019}. In such methods, the objective function is designed to track the operator's desired nominal command~\cite{fundaConstrainedCartesianMotion1996}, which can be obtained from a force sensor or from a master robot~\cite{kapoorConstrainedControlSurgical2006}. Early frameworks included active constraints in both the objective and the optimization constraints --- however, mixing active constraint and command-tracking terms in the objective makes their relative weights difficult to tune~\cite{kapoorConstrainedControlSurgical2006}. Later work avoids this problem by expressing active constraints purely as optimization constraints, reserving the objective for command tracking~\cite{marinhoUnifiedFrameworkTeleoperation2019}.

In contrast, handheld systems cannot implement force-based methods due to the lack of a mechanical ground separate from the operator. This necessitates modified formulations where the handheld robot actuates/deflects independently of the handle to satisfy the active constraints~\cite{beckerHandheldMicromanipulationVisionbased2011}. Given a feature that the tool tip wants to be constrained with respect to, e.g., a point, line, or plane, the formulation in~\cite{beckerVisionBasedControlHandheld2013} constrains the tip to the closest point on the constraint feature, but is unable to constrain multiple points and/or orientations of the robot. In the case of multiple constraint features, operating close to feature boundaries may cause undesirable switching behavior, which~\cite{beckerVisionBasedControlHandheld2013} handles by disabling all constraints except for the closest one. For safety-critical applications like microsurgery, this means that enforcing safety-related constraints prevents guidance-related constraints from being enabled at the same time.

Active constraints implemented using constrained optimization can address the above limitations, but existing formulations for grounded systems cannot be generalized to handheld systems. This stems from two main differences: handheld robots have much smaller workspaces due to their size, and actuate relative to the operator's hand. Because the operator realizes the desired command with their own hand, the command-tracking objective term~\cite{marinhoUnifiedFrameworkTeleoperation2019} is degenerate. Additionally, hand motion is an external input that the controller cannot measure in advance or modify, so active constraints can only be enforced against predictions, in which satisfaction only holds while the required enforcement is within the robot workspace.

\section{Optimization-based Active Constraint Framework for Handheld Micromanipulation}
\label{sec:methods}
\subsection{System Model}
In this formulation, we consider a handheld robot that can actuate in the six rigid body degrees of freedom (6-DoF) relative to the handle as in~\cite{yangDesignAnalysis62012}. We also define three reference frames: an arbitrary task-space frame, a handle frame rigidly attached to the instrument handle, and a robot frame rigidly attached to the output of the 6-DoF robot. Unless indicated otherwise, all quantities are expressed in task-frame coordinates. The robot pose $T_k \in \mathrm{SE}(3)$ at timestep $k \in \mathbb{Z}_{\ge 0}$, i.e., the pose of the robot frame expressed in task-frame coordinates, can then be written as a discrete-time kinematic update
\begin{equation}
    \label{eq:kinematic_model}
    T_{k+1} = \Delta T^{\mathrm{robot}}_{k} \cdot \Delta T^{\mathrm{hand}}_{k} \cdot T_k
\end{equation}
where $\Delta T^{\mathrm{robot}}_{k} ,\, \Delta T^{\mathrm{hand}}_{k} \in \mathrm{SE}(3)$ are the pose increments contributed over the timestep by the robot and hand, respectively. At timestep $k$, the controller commands $\Delta T^{\mathrm{robot}}_{k}$ but the hand motion $\Delta T^{\mathrm{hand}}_{k}$ is not known ahead of time, so the controller works with the estimate $\Delta \hat{T}^{\mathrm{hand}}_{k}$. We define \textit{drift pose} $\hat{T}^{\mathrm{drift}}_{k+1} \coloneqq \Delta \hat{T}^{\mathrm{hand}}_{k}\cdot T_k$ as the predicted robot pose that would result if no actuation is commanded over the timestep. To avoid optimizing over manifold constraints, we define optimization variable $\boldsymbol{\xi} \coloneqq [\Delta \mathbf{p}^\top, \Delta \boldsymbol{\theta}^\top]^\top \in \mathbb{R}^6$ as the local tangent-space coordinates of $\Delta T^{\mathrm{robot}}_{k}$ with translation referenced at a point $\mathbf{p}$ on the robot, recovered through the exponential map and the spatial adjoint of $\mathbf{p}$. Here $\Delta \mathbf{p} \in \mathbb{R}^3$ is the displacement of $\mathbf{p}$ over the timestep, while $\Delta \boldsymbol{\theta} \in \mathbb{R}^3$ is the corresponding orientation increment as a rotation vector.

\subsection{Optimization Structure}
We build the framework on quadratic programming (QP) as in prior works~\cite{marinhoDynamicActiveConstraints2019a}, such that enforcing active constraints at each timestep equates to solving a QP of the form
\begin{equation}
    \label{eq:qp_standard}
    \boldsymbol{\xi}^*=\argmin_{\boldsymbol{\xi}}\ \tfrac{1}{2}\boldsymbol{\xi}^\top \boldsymbol{H} \boldsymbol{\xi} + \boldsymbol{g}^\top \boldsymbol{\xi}
    \quad \, \mathrm{s.t.} \quad \, \boldsymbol{l} \le \boldsymbol{C}\boldsymbol{\xi} \le \boldsymbol{u}
\end{equation}
for the optimal command $\boldsymbol{\xi}^*$, where the objective Hessian $\boldsymbol{H} \succ 0$ and the objective gradient $\boldsymbol{g}$ are derived from the objective terms detailed in Section~\ref{subsec:objective}, while $\boldsymbol{C}$, $\boldsymbol{l}$ and $\boldsymbol{u}$ define the optimization constraints detailed in Section~\ref{subsec:constraints}.

Here, safety (active) constraints are enforced as optimization constraints, while guidance (active) constraints are expressed in the objective. While recent grounded formulations reserve the objective for command tracking and express all active constraints as optimization constraints to avoid balancing multiple objectives and their weights~\cite{marinhoDynamicActiveConstraints2019a}, handheld systems do not suffer from the same problem due to the lack of a command-tracking term.

\subsection{Active Constraint Primitives}
\label{subsec:primitives}
Desired active constraint behavior can be formed using \textit{constraint primitives}. A constraint primitive defines a desired relationship between a \textit{primitive feature} on the robot and the environment. A primitive feature can be a point, line (curve), or plane (surface), while the type of relationship can be some form of Euclidean or angular distance. The polarity of the relationship, which may be keep-in or keep-out, also needs to be specified. For instance, constraining a robot point within some distance of an environmental point calls on the point--point distance keep-in primitive, which forms a spherical allowable region. We define scalar function $\phi$ as the \textit{constraint value}
\begin{equation}
    \label{eq:generic_primitive}
    \phi(T) = \pm \left( \rho \left(\mathcal{F}_\mathrm{robot}\left(T \right),\, \mathcal{F}_\mathrm{env} \right) - \rho_0 \right)
\end{equation}
which encodes constraint primitives, such that $\phi \ge 0$ when the constraint is satisfied, and $\phi < 0$ otherwise. A continuous function $\rho(\cdot, \cdot)$ returns a scalar quantifying the Euclidean or angular separation (e.g., 2-norm, geodesic distance) between environment feature $\mathcal{F}_\mathrm{env}$ and robot feature $\mathcal{F}_\mathrm{robot}$, the latter of which is a function of robot pose $T$, while scalar $\rho_0 \ge 0$ sets the threshold of satisfaction. The prefixed sign sets the constraint polarity: positive for keep-out, where $\rho \ge \rho_0$ is required, and negative for keep-in, where $\rho \le \rho_0$ is required. Combining constraint primitives forms a compound active constraint, which is the intersection of the superlevel sets formed by the constituent constraint primitives. These constraint primitives are similar to those in~\cite{fundaConstrainedCartesianMotion1996,kapoorConstrainedControlSurgical2006,marinhoDynamicActiveConstraints2019a}, which contain derivations of instantiated primitives.

The distance function $\rho$, and consequently the constraint value $\phi$, is generally nonlinear in the optimization variable $\boldsymbol{\xi}$. We therefore expand the predicted constraint value at the next timestep $\hat{\phi}(T_{k+1})$ about the drift pose to first order, assuming environmental features are static over the timestep:
\begin{equation}
    \label{eq:linearization}
    \hat{\phi}(T_{k+1}) \, \approx \, \hat{\phi}^{\mathrm{drift}}_{k+1} \, + \, \mathbf{a}_k^\top \boldsymbol{\xi} ,
\end{equation}
where $\hat{\phi}^{\mathrm{drift}}_{k+1} \coloneqq \phi(\hat{T}^{\mathrm{drift}}_{k+1})$ is the constraint value of the next timestep that would result from predicted hand motion only, and $\mathbf{a}_k \in \mathbb{R}^6$ is the gradient of $\phi$ with respect to $\boldsymbol{\xi}$ evaluated at the drift pose. This decomposition separates the constraint value into what the hand contributes and what the commanded robot pose increment can add.

\subsection{Objective Function}
\label{subsec:objective}
Because handheld systems lack the command-tracking objective term that grounded systems require, the optimization objective can carry guidance constraints without the difficulty of balancing multiple primary objectives~\cite{kapoorConstrainedControlSurgical2006}. In addition, we append objective terms that encode the secondary objective of minimizing total actuator cost (e.g., stroke consumption) in enforcing active constraints, which resolves the optimization solution in the null space of the active constraints.

\subsubsection{Actuator Cost}
To minimize total actuator cost, we append a term penalizing the magnitude of the commanded increment (referred to as the \textit{rate term}), as well as a term penalizing the total excursion from the home configuration where actuator cost is zero (referred to as the \textit{home term}). The penalty is defined in actuator space to account for the variable mapping between task space and actuator space, which depends on hand position and robot configuration. We therefore write the corresponding actuator increment as $\Delta \mathbf{q}_k \approx \boldsymbol{B}_k \boldsymbol{\xi}$, where $\boldsymbol{B}_k$ composes the change of frame from expressing $\boldsymbol{\xi}$ in the handle frame (which the robot actuates relative to) with the inverse kinematic Jacobian of the robot.
The two penalty terms together are then
\begin{equation}
    \label{eq:actuator_term}
    w^{\mathrm{r}} \norm{\boldsymbol{B}_k \, \boldsymbol{\xi}}^2
    \quad + \quad w^{\mathrm{h}} \norm{\boldsymbol{B}_k \, \boldsymbol{\xi} + \mathbf{q}_k}^2
\end{equation}
with scalar weights $w^{\mathrm{r}}$ and $w^{\mathrm{h}}$ for the rate and home terms, respectively, while $\mathbf{q}_k$ denotes the actuator displacements from home configuration at the current timestep. Because $\boldsymbol{B}_k$ already encodes the anisotropy of actuator costs in different task space directions, scalar weights are used for ease of parameter tuning. The rate and home weights can be respectively interpreted as derivative and proportional gains that define the robot's homeward speed.

\subsubsection{Guidance Constraints}
Guidance-related active constraints enter the optimization objective as a quadratic penalty on the scalar (guidance) constraint value:
\begin{equation}
    \label{eq:guidance_term}
    w^{\mathrm{g}}(e_k \, + \, \mathbf{a}_k^\top \boldsymbol{\xi})^2
\end{equation}
with $e_k \coloneqq \hat{\phi}^{\mathrm{drift}}_{k+1} - \phi^\mathrm{set}_k$ denoting the error between the guidance constraint value at the drift pose $\hat{\phi}^{\mathrm{drift}}_{k+1}$ and guidance constraint value setpoint $\phi^\mathrm{set}_k$. Generally, guiding the robot along a trajectory calls for a keep-in constraint primitive, where the constraint threshold $\rho_0$ from \eqref{eq:generic_primitive} is typically zero, in which case the constraint value $\phi$ simplifies to the negated distance function $-\rho$. Correspondingly, the constraint value setpoint is also typically zero as it is desirable for the distance between the robot and the trajectory to be zero, in which case the guidance term in \eqref{eq:guidance_term} penalizes the robot's distance from the trajectory. At each timestep, the commanded increment $\boldsymbol{\xi}$ corrects a fraction of $e_k$, subject to the actuator cost terms in \eqref{eq:actuator_term}, such that the constraint value is driven towards setpoint $\phi^\mathrm{set}_k$ over time.
\subsubsection{Guidance Weight Normalization}
\label{subsec:guidance_bandwidth}
Generally, relative weights in optimization-based frameworks are hard to predict and must be retuned as the system or task changes~\cite{kapoorConstrainedControlSurgical2006}. Here, a similar problem arises for a different reason: the fraction of guidance constraint value error $e_k$ corrected per timestep (i.e., bandwidth of guidance) depends on the actuator cost along the constraint gradient $\mathbf{a}_k$, i.e., the actuator cost of providing guidance. This cost can vary significantly for handheld robots, even with fixed weights on all objective terms; consider an example of a guidance constraint that pins the tip of a long end effector (extended from the robot) at a fixed target point. Given some small initial distance between the tip and the target point, there are combinations of hand pose and robot configuration which can utilize the moment arm of the long end effector to close the tip--target distance such that the robot has to actuate minimally, while there are other combinations where the end effector's moment arm cannot be utilized to close the distance such that the robot has to actuate maximally. Because the actuator cost of providing guidance varies significantly under fixed weights, so does the bandwidth or ``amount'' of provided guidance, which is generally undesirable. We address this issue by normalizing the guidance weight against actuator cost along constraint gradient $\mathbf{a}_k$ at each timestep using
\begin{equation}
    \label{eq:normalization}
    w^{\mathrm{g}}_k \,=\, \frac{\mu}{2 \, \mathbf{a}_k^\top \boldsymbol{H}^{-1}_0 \mathbf{a}_k}
\end{equation}
where $\mu$ sets the dimensionless ratio, while $\boldsymbol{H}_0$ denotes the actuator terms' accumulated contributions to the objective Hessian in \eqref{eq:qp_standard} such that $\boldsymbol{H} = \boldsymbol{H}_0 + 2\,w^{\mathrm{g}}_k \, \mathbf{a}_k \, \mathbf{a}_k^\top$. Here $\mathbf{a}_k^\top \boldsymbol{H}^{-1}_0 \mathbf{a}_k$ represents the compliance of the actuator terms along the constraint gradient, which is large when guidance is cheap to provide and small when it is expensive. Accordingly, the dynamics of the constraint value error $e_k$ (in the absence of active optimization constraints) can be derived by projecting the stationarity condition of \eqref{eq:qp_standard} onto $\mathbf{a}_k$ to get
\begin{equation}
    \label{eq:error_dynamics}
    e_{k+1} \,=\, \frac{1}{1 + \mu}\,e_k \; - \; \frac{G_0}{1 + \mu}
\end{equation}
where $G_0 \coloneqq \mathbf{a}^\top_k \boldsymbol{H}^{-1}_0 \boldsymbol{g}_0$ is an offset term resulting from normalization, representing the change in constraint value in the absence of the guidance term, with $\boldsymbol{g}_0$ denoting the objective gradient contribution from the home term in the form of \eqref{eq:qp_standard}. This offset imposes a steady-state error of $-G_0/\mu$ which can be compensated for by shifting the setpoint accordingly. As \eqref{eq:error_dynamics} is a linear time-invariant difference equation, it can be characterized by a first-order pole with cutoff frequency $f_c \approx \ln(1+\mu)/(2\pi\Delta t)$ for controller loop period $\Delta t$, such that $\mu$ sets the guidance bandwidth invariant to parameters and weights. In addition to simplifying parameter tuning, the bandwidth property is also desirable when the disturbances that the guidance constraints act against are band-limited (e.g., hand tremor), in which case the guidance weight no longer needs to be set to a large value to maintain consistent behavior, avoiding numerical ill-conditioning of the QP.

\subsubsection{Guidance Degradation}
\label{subsec:guidance_degradation}
Due to the limited actuation budget (i.e., the available actuator stroke) of handheld robots, it is not always possible to enforce all guidance and safety active constraints within the robot's workspace. In this case, guidance should degrade in a perceptible manner so that the operator can correct their hand position in a way which recovers stroke. This could be done by scaling the guidance weight $w^{\mathrm{g}}_k$, but the guidance bandwidth discussed in Section~\ref{subsec:guidance_bandwidth} would also change accordingly. Because the robot acts within the operator's eye--hand feedback loop, which has a critical frequency of \SIrange{0.5}{2}{\hertz}, varying the frequency response of assistance risks destabilizing effects~\cite{maclachlanMicronActivelyStabilized2012a}. Instead, we vary the constraint value setpoint $\phi^\mathrm{set}_k$ using
\begin{equation}
    \label{eq:guidance_degradation}
    \phi^\mathrm{set}_k = (1 \,-\, \alpha_k) \, \phi^\mathrm{off}_k
\end{equation}
where $\alpha_k \in [0,1]$ is the \textit{guidance fraction} that blends the nominal constraint value setpoint of zero (full guidance) with the guidance-off constraint value $\phi^\mathrm{off}_k \coloneqq \phi(T^\mathrm{off}_k)$, which is defined as the guidance constraint value evaluated at the minimum-actuator-cost robot pose $T^\mathrm{off}_k$ that satisfies all safety constraints and joint/actuator limits while ignoring guidance. In other words, full guidance is provided when $\alpha_k = 1$, degrading towards no guidance as $\alpha_k \to 0$, all while safety is enforced. Guidance-off pose $T^\mathrm{off}_k$ is obtained by sequentially solving the same QP without the guidance term, starting from the home configuration and re-evaluating/re-linearizing optimization constraints (containing safety constraints and joint/actuator limits) with each solve until optimization constraint residuals fall below a tolerance.

Guidance fraction $\alpha_k$ is derived from the actuator stroke needed to provide full guidance while enforcing safety. We estimate this stroke to first order by solving the underdetermined system $\mathbf{a}^\top_k \boldsymbol{\xi}= -\phi^\mathrm{off}_k$ which has a closed form minimum-norm solution $\boldsymbol{\xi}^\mathrm{on}_k = -\phi^\mathrm{off}_k \, \mathbf{a}_k \,/ \norm{\mathbf{a}_k}^2$ representing the minimum-norm pose increment from guidance-off pose $T^\mathrm{off}_k$ that approximately drives the guidance-off constraint value $\phi^\mathrm{off}_k$ to zero. The stroke required to provide full guidance and safety is then defined as $\mathbf{q}^\mathrm{on}_k \coloneqq \mathbf{q}^\mathrm{off}_k + \boldsymbol{B}_k \boldsymbol{\xi}^\mathrm{on}_k$, where $\mathbf{q}^\mathrm{off}_k$ denotes the actuator displacements corresponding to guidance-off pose $T^\mathrm{off}_k$ from inverse kinematics. Worst-case stroke fraction $s_k$ and guidance fraction $\alpha_k$ follow as
\begin{equation}
    \label{eq:guidance_fraction}
    s_k \coloneqq \frac{\norm{\mathbf{q}^\mathrm{on}_k}_\infty}{q^\mathrm{max}} ,
    \qquad
    \alpha_k = \mathrm{clamp}\!\left(\frac{s^\mathrm{off} - s_k}{s^\mathrm{off} - s^\mathrm{on}},\, 0,\, 1\right)
\end{equation}
where $q^\mathrm{max}$ denotes the maximum allowable actuator stroke, which we assume here to be uniform across actuators, while $s^\mathrm{on}$ and $s^\mathrm{off}$ demarcate the degradation boundaries on $s_k$ such that guidance is fully provided below $s^\mathrm{on}$ and fully disabled above $s^\mathrm{off}$. Because $\mathbf{q}^\mathrm{on}_k$ accounts for the stroke required to enforce safety, in the case that providing guidance recovers stroke, $s_k$ stays small and guidance is not degraded.

\subsection{Optimization Constraints}
\label{subsec:constraints}
\subsubsection{Mechanism Limits}
Static and dynamic limits of the robot joints and actuators are encoded in the optimization constraints to prevent unrealizable solutions. As these limits are typically nonlinear in the decision variable $\boldsymbol{\xi}$, we relinearize the constraints at each timestep before the QP solve.
\subsubsection{Safety Constraints}
\label{subsec:safety}
For a safe set $\mathcal{S}$ of robot poses defined as the superlevel set of continuous scalar function $\phi$, i.e., $\mathcal{S}=\{T \in \mathrm{SE}(3) : \phi(T) \ge 0\}$, enforcing the discrete control barrier function (DCBF) condition $\phi(T_{k+1}) \ge (1-\gamma)\phi(T_k)$ with barrier gain $\gamma \in (0, 1]$ (which we note is similar to vector field inequalities described in~\cite{marinhoDynamicActiveConstraints2019a}) guarantees forward invariance of the safe set $\mathcal{S}$ if $\phi(T_0) \ge 0$~\cite{agrawalDiscreteControlBarrier2017}. Since $\phi(T_{k+1})$ is not known at solve time, we substitute the first order expansion of the prediction \eqref{eq:linearization}. However, errors from linearization, hand motion prediction, imperfect command tracking, etc., induce discrepancies between the actual and predicted constraint value. Following robust CBF formulations~\cite{jankovicRobustControlBarrier2018}, we absorb these errors into a prespecified safety margin $m \ge 0$ and enforce the DCBF condition on the tightened set $\{T : \phi(T) \ge m \}$. Forward invariance therefore holds while the prediction residual is within $m$~\cite{kolathayaInputtoStateSafetyControl2019}.

Additionally, enforcing safety constraints may demand motion that the mechanism limits prevent, leaving the QP infeasible. We therefore soften the safety constraints with slack terms $\sigma$ and augment the decision vector and objective function accordingly, with the slack term penalty weights set well above the objective weights introduced in Section~\ref{subsec:objective} so the slack terms go nonzero only when no feasible solution exists. Slack terms are added to the safety constraints rather than the mechanism limits to prevent commands that may damage the robot --- if both sets of constraints are binding, safety cannot be enforced regardless of where the slack terms are placed. This gives
\begin{equation}
    \label{eq:safety_inequality}
    \mathbf{a}_{k}^\top \boldsymbol{\xi} \,+\, \sigma \ge\ -\gamma\left(\phi_k - m\right) - \Delta \hat{\phi}^\mathrm{hand}_k
\end{equation}
for a single safety constraint primitive, where $\phi_k \coloneqq \phi(T_k)$ is the constraint value evaluated at current robot pose $T_k$, while $\Delta \hat{\phi}^\mathrm{hand}_k \coloneqq \hat{\phi}^{\mathrm{drift}}_{k+1} \,-\, \phi_{k}$ denotes the constraint value increment contributed by predicted hand motion over the timestep, so safety is enforced in a feedforward manner --- this term is not present for grounded systems, where a zero (do-nothing) command maintains safety.

\section{Experiments\protect\footnotemark}
\footnotetext{See accompanying video that demonstrates the system and the task}
We evaluated the framework on a handheld robot performing a medically relevant artificial task that requires guidance and safety enforcement at the same time.

\subsection{Task Description}
The artificial task was defined with a performance goal and a safety requirement to demonstrate guidance provision and safety enforcement of the proposed active constraint formulation: the goal was to hold a needle tip at a target point for \SI{330}{\second} under a microscope while satisfying a remote center of motion (RCM) safety requirement that lay outside the microscope field of view. The long-duration needle tip hold-still is inspired by retinal vein cannulation~\cite{gonencRobotassistedRetinalVein2016}, a vitreoretinal microsurgical procedure, whereas the RCM constraint corresponds to the instrument port, which is also typically outside the microscope view. 

\begin{figure}[t]
    \centering
    \includegraphics[width=\columnwidth, trim=15mm 87mm 20mm 80mm, clip=true]{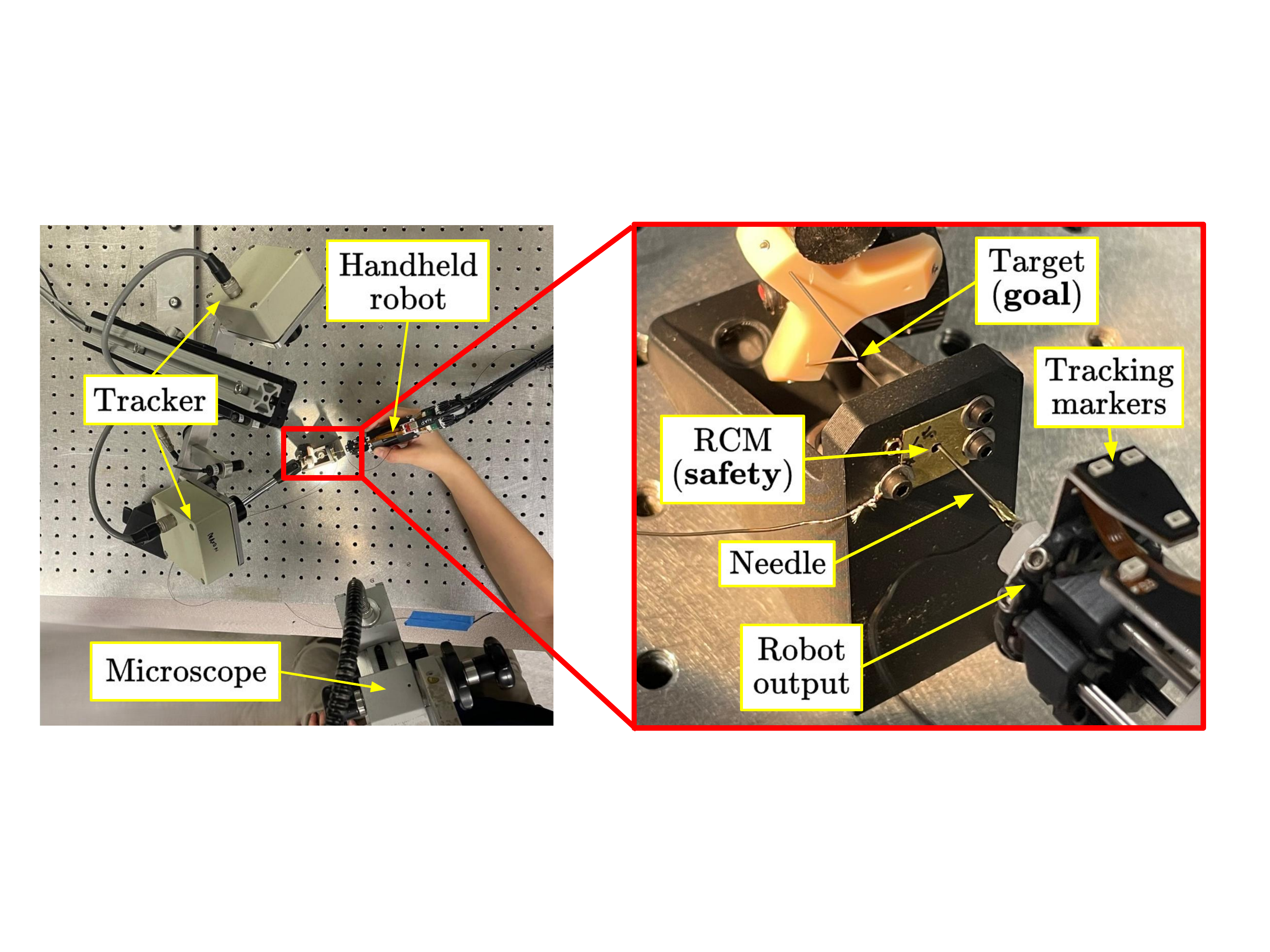}
    \caption{Experimental setup of the artificial task, where the goal was to hold the needle tip at the target point while satisfying a remote center of motion (RCM) safety constraint. The target point was a virtual point defined by the intersection of two hypodermic needles, while the RCM was realized by a small hole in a shim.}
    \label{fig:fig1_setup}
\end{figure}

The RCM was realized as a hole in a brass shim, through which the needle was inserted (Fig.~\ref{fig:fig1_setup}). To minimize plastic deformation, the needle was fabricated from superelastic nitinol wire with a nominal diameter of \SI{610}{\micro\meter}, which is comparable to incision diameters of vitreoretinal microsurgery~\cite{mohamedReviewSmallGauge2017}. The target point was a virtual point defined by the intersection of two 24-gauge hypodermic needles, and was positioned approximately \SI{20}{\milli\meter} beyond the RCM (in the direction of needle insertion) to approximate intraocular working distances~\cite{yangTechniquesRobotaidedIntraocular2018}. For a shim thickness of \SI{160}{\micro\meter}, the hole diameter was nominally set to \SI{1320}{\micro\meter}, approximately twice the needle diameter, so that the task remained feasible unaided. Contact between the needle shaft and the hole edge was deemed a safety violation, and was recorded with electrical continuity at \SI{1}{\kilo\hertz}. Since the microscope shows only the needle tip, the operator cannot observe the shaft--hole clearance and must infer contact from tip motion.
\subsection{Implementation Details}
The proposed framework was implemented on a 6-DoF parallel kinematic handheld robot (micromanipulator) with a translational workspace that circumscribes a cylinder \SI{4}{\milli\meter} in diameter and height, along with a rotational workspace of $\pm \SI{7.5}{\degree}$ about the three axes of rotation, each measured from the home configuration. An optical tracker~\cite{maclachlanHighSpeedMicroscaleOptical2009} measured the robot and handle poses in its own coordinate frame (defined as the task frame) at a rate of \SI{1}{\kilo\hertz} with a noise floor of approximately \SI{10}{\micro\meter} RMS. The needle tip and shaft axis were registered to the robot frame using the tracker and least-squares fitting, while the target point and RCM point were similarly registered to the tracker (task) frame. Following~\cite{beckerVisionBasedControlHandheld2013}, hand motion was predicted by integrating forward the estimated velocity from a constant-acceleration error-state Kalman filter~\cite{solaMicroLieTheory2018} on the tracked handle pose.

The controller was implemented in ROS2 Jazzy Jalisco on Ubuntu 24.04 with the real-time kernel patch, in which the \SI{1}{\kilo\hertz} control loop was pinned to an isolated core of an Intel Core Ultra 7 270K Plus. Both optimization programs discussed in Section~\ref{sec:methods} were solved with ProxQP~\cite{bambadePROXQPAnotherQuadratic2022}, giving mean and median total solve times of \SI{27}{\micro\second} and \SI{24}{\micro\second}.

Two distance-based primitives of the form \eqref{eq:generic_primitive} were instantiated (with feature pairs denoted as robot feature--environmental feature): a point--point keep-in primitive as the guidance constraint for holding the needle tip at the target point, and a line--point keep-in primitive as the safety constraint for restricting needle axis displacement from the hole center. While other formulations like~\cite{marinhoDynamicActiveConstraints2019a} used the squared 2-norm as the distance function $\rho$ to avoid derivative singularities of the standard 2-norm, we used the Huber norm~\cite{huberRobustEstimationLocation1992} here, which addresses the same problem while retaining linear distance units (outside the transition region) to simplify parameter tuning. In the experiments, the Huber transition parameter $\delta$ was set below the noise floor of the tracker to minimize effects of the quadratic regime. 

The guidance constraint used $\mu = 0.25$, corresponding to an approximate bandwidth of \SI{36}{\hertz} per \eqref{eq:error_dynamics}, which was limited by overall system bandwidth and delay, but still above the dominant physiological hand tremor band of \SIrange{4}{12}{\hertz}~\cite{lenkaTremorSyndromesUpdated2021} and well above the human eye--hand feedback loop's critical frequency of \SIrange{0.5}{2}{\hertz}~\cite{maclachlanMicronActivelyStabilized2012a}. Guidance degradation boundaries were $s^\mathrm{on}=0.50$ and $s^\mathrm{off}=0.85$ to prevent full actuator saturation. The safety constraint used distance threshold $\rho_0 = \SI{355}{\micro\meter}$, equal to the nominal shaft--hole clearance radius, while the safety margin was set to $m = \SI{200}{\micro\meter}$, informed by registration residuals and estimated servo error. The barrier gain $\gamma = 0.05$ was kept low to avoid oscillatory behavior at the constraint boundary caused by overall system delay.

\begin{figure*}[t]
    \centering
    \includegraphics[width=\textwidth, trim=0mm 0mm 0mm 0mm, clip=true]{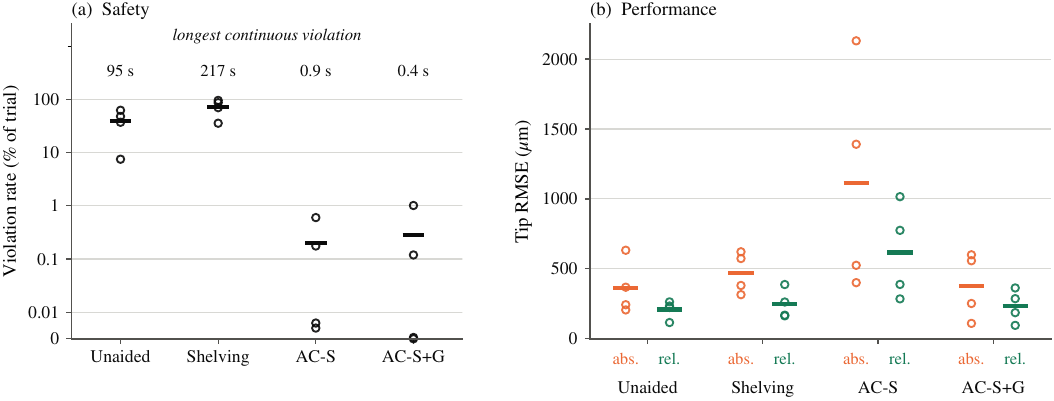}
    \caption{Safety and task performance over four subjects. Each marker is a subject; the horizontal rule gives the mean over subjects. Each subject's two repeated trials were pooled before averaging. (a) Violation rate, with a logarithmic axis above \SI{0.01}{\percent} and linear below so that zero is shown; longest continuous violation per condition is printed above. (b) Tip RMSE against the target (abs.) and against the trial mean tip position (rel.) from contact-free samples. The proposed AC-S+G lowers violation rate by more than two orders of magnitude against Unaided, and recovers the accuracy that AC-S loses.}
    \label{fig:fig2_results}
\end{figure*}
\subsection{Conditions and Protocol}
The aforementioned task was performed under four conditions: 1) \emph{Unaided}, with the robot disabled; 2) \emph{Shelving}, an established task-agnostic assistance algorithm for handheld robots~\cite{maclachlanMicronActivelyStabilized2012a} that attenuates hand tremor; 3) \emph{AC-S}, the proposed active constraint framework enforcing safety only; and 4) \emph{AC-S+G}, the complete proposed framework with safety and guidance both enabled. Unaided serves as the baseline, while comparing Shelving against the latter two conditions isolates the proposed active constraints on identical hardware. Finally, comparing AC-S against AC-S+G quantifies the cost of enforcing safety without guidance.

Four novice subjects completed two trials per condition in a pilot study under a board-approved protocol. Before the trials, each subject trained on the task unaided to adapt to manipulation at a small scale under the microscope, and to develop the intuition of detecting safety violations from tip motion. They were then introduced to each of the three assisted conditions for a similar duration. Condition order was assigned by a balanced Latin square~\cite{williamsExperimentalDesignsBalanced1949}, with the second set of trials reversing the order of the first to mitigate ordering and fatigue effects.

\subsection{Evaluation Metrics}
The first \SI{30}{\second} of each trial were treated as an acclimatization period and excluded from analysis, leaving \SI{300}{\second} per trial, so that the reported metrics reflect steady-state performance rather than the initial approach and settling. 

Task performance was evaluated using the root mean square of the Euclidean distance between tip position and target point, $\mathrm{RMSE}_\mathrm{abs}$. To account for the fact that the subjects may not perceive the same target as the registered point, we also report $\mathrm{RMSE}_\mathrm{rel}$ for comparisons between conditions, defined as the $\mathrm{RMSE}$ between tip position and mean tip position over the trial period. Because tip position is computed from measured robot pose and a calibrated tip offset, we compute these performance metrics only from samples where continuity was open, as contact undermines the implicit rigid-body assumption. Within each subject and condition, performance metrics across the two repeats were combined by root mean square, weighted by the number of contact-free samples in each, while each condition was summarized by the arithmetic mean across subjects.

Correspondingly, we quantify task safety with \textit{violation rate}, defined as the percentage of trial time where continuity was closed, averaged across all trials of the same condition. As a worst-case measure, we also report the longest continuous violation over all trials of the same condition.

\section{Results and Discussion}
\subsection{Safety}
\label{subsec:safety_results}
Figure~\ref{fig:fig2_results}(a) shows the safety metrics across conditions. In the two baseline conditions, Unaided and Shelving, safety was violated for a substantial portion of each trial: \SI{38.7}{\percent} and \SI{72.0}{\percent}, respectively. Enforcing the safety constraint reduced average violation rate to \SI{0.2}{\percent} and \SI{0.3}{\percent} for AC-S and AC-S+G, respectively --- a reduction of more than two orders of magnitude against both baselines, which held for all four subjects. The worst-case measure of the longest continuous violation similarly shows the same separation, going from \SI{95}{\second} and \SI{217}{\second} for the Unaided and Shelving condition respectively, to \SI{0.9}{\second} for AC-S and \SI{0.4}{\second} for AC-S+G. Most importantly, activating guidance in AC-S+G did not jeopardize safety, as AC-S and AC-S+G did not differ meaningfully on either safety metric.

Violation rate was higher for all subjects under Shelving compared to Unaided. Although the Shelving assistance algorithm attenuates operator hand tremor, it does so without knowledge of the safety requirements such that the provided assistance can compromise safety. Furthermore, since the subjects developed intuition for the task under the Unaided condition, the lag introduced by the Shelving algorithm's low-pass filter~\cite{maclachlanMicronActivelyStabilized2012a} may have impaired the skill transfer. Specifically, because the tip motion cue for detecting safety violations is altered under the Shelving algorithm, the subjects may have had more difficulty identifying shaft--hole contact and correcting appropriately.

\begin{figure*}[t]
    \centering
    \includegraphics[width=\textwidth, trim=0mm 0mm 0mm 0mm, clip=true]{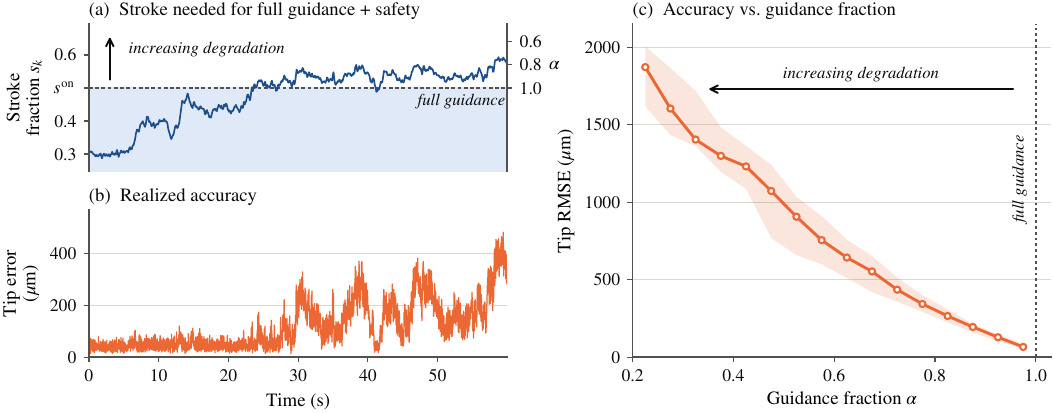}
    \caption{Guidance degrading to preserve safety, due to insufficient actuation budget. Tip error is the distance from the tip to the registered target point. (a) and (b) show an excerpt of an AC-S+G trial, as means over \SI{20}{\milli\second} windows. (a) Stroke fraction $s_k$ and the corresponding guidance fraction $\alpha_k$. Below $s^\mathrm{on}$, guidance is fully provided ($\alpha_k = 1$), degrading when $s_k > s^\mathrm{on}$. (b) Corresponding tip error, which approximately tracks $(1-\alpha_k)$ proportionally. No safety violation occurred in this window. (c) Tip RMSE against the guidance fraction $\alpha_k$ over all eight AC-S+G trials and binned in uniform intervals with a width of 0.05; the line is computed from the pooled data while the band spans the unpooled RMSE range across trials.}
    \label{fig:fig3_degradation}
\end{figure*}

\subsection{Task Performance}
Figure~\ref{fig:fig2_results}(b) shows the task performance metrics across conditions. Averaged across subjects, the Unaided condition produced $\mathrm{RMSE}_\mathrm{abs}$ of \SI{363}{\micro\meter}, with a minor increase to \SI{473}{\micro\meter} under the Shelving condition. Compared to the Unaided condition, enforcing only safety increased $\mathrm{RMSE}_\mathrm{abs}$ for all four subjects, resulting in a 3$\times$ increase to \SI{1112}{\micro\meter}. Enabling guidance on top of enforcing safety in the AC-S+G condition produced a 3$\times$ decrease (from AC-S) to \SI{380}{\micro\meter}, compensating for the decrease in accuracy from just enforcing safety. $\mathrm{RMSE}_\mathrm{rel}$ shows the same pattern: \SI{210}{\micro\meter} in the Unaided case increased slightly to \SI{246}{\micro\meter} in the Shelving case, while enforcing only safety in AC-S resulted in a much larger $\mathrm{RMSE}_\mathrm{rel}$ of \SI{616}{\micro\meter}, counteracted by enabling guidance in AC-S+G to give \SI{233}{\micro\meter}. Because both performance metrics agree in the relative effects of each condition, the results suggest that the problem of subjects perceiving a different target point is not a concern.

The minor reduction in accuracy under the Shelving assistance can be explained by a similar reason to Section~\ref{subsec:safety_results}, where the assistance algorithm may have undermined the task intuition developed in the Unaided condition, which is exacerbated by the cognitive load imposed by the safety requirement of avoiding shaft--hole contact. On the other hand, the accuracy loss under the safety-only AC-S condition agrees with the geometry of the task. Without a guidance constraint holding the needle tip at the target point, the minimum-actuator-cost robot pose that keeps the shaft clear of the hole edge in this case moves the tip away from the target. As the line--point RCM safety constraint only restricts two degrees of freedom, the addition of a guidance constraint in the AC-S+G condition enables the optimization to use the remaining degrees of freedom of the robot to satisfy both active constraints. AC-S is also representative of what a single-constraint framework like~\cite{beckerVisionBasedControlHandheld2013} can do in this task, as such a framework must disable guidance whenever the safety constraint is binding. AC-S approximates the worst case of this, with guidance disabled throughout, resulting in an accuracy loss of 3$\times$.

Furthermore, the minimum-actuator-cost poses that enforce the RCM safety constraint in this case exhibit the fulcrum effect, i.e., tip motion is reversed relative to the hand's motion, which may be unintuitive to a novice operator without (micro)surgical experience. Similar to the Shelving assistance, the task intuition developed in the Unaided condition may not transfer well as a result. This phenomenon also introduces skill as a confounding factor, where subjects that intuited the fulcrum effect could have better utilized the assistance in the two active constraint (AC) conditions, which possibly explains the higher variability seen in Fig.~\ref{fig:fig2_results}.

\subsection{Guidance Degradation}
Figure~\ref{fig:fig3_degradation}(a) and (b) show the guidance degradation mechanism during a \SI{60}{\second} window of an AC-S+G trial. Tip error stays below \SI{100}{\micro\meter} while stroke fraction $s_k$ remains under degradation boundary $s^\mathrm{on}$, where guidance is fully provided. As $s_k$ exceeds $s^\mathrm{on}$ and guidance fraction $\alpha_k$ decreases, the tip error approximately tracks the degraded setpoint $(1-\alpha_k)\phi^\mathrm{off}_k$ of \eqref{eq:guidance_degradation}. Figure~\ref{fig:fig3_degradation}(c) shows the same relation over all eight AC-S+G trials, where tip RMSE increases monotonically as the guidance fraction $\alpha_k$ decreases to preserve safety under limited actuation budget. The tip RMSE is \SI{64}{\micro\meter} for samples under full guidance ($\alpha_k = 1$), against \SI{486}{\micro\meter} over the degraded samples ($\alpha_k < 1$) and \SI{434}{\micro\meter} overall, pooled across trials rather than averaged across subjects as in Fig.~\ref{fig:fig2_results}.

The guidance degradation mechanism was exercised over most of its range. Degradation was active for \SI{79}{\percent} of all samples ($\alpha_k < 1$), and the largest observed stroke fraction $s_k$ of 0.78 corresponds to guidance fraction $\alpha_k = 0.20$ per \eqref{eq:guidance_fraction}, meaning guidance was never fully disabled and the actuation budget was never fully depleted.

The distribution of $s_k$ indicates that the subjects used guidance degradation as a cue for making corrections in hand positioning, which was a design objective introduced in Section~\ref{subsec:guidance_degradation}: $s_k \in [0.50, 0.55)$ occupied \SI{35.5}{\percent} of all samples while $[0.45, 0.50)$ occupied only \SI{4.5}{\percent}. These adjacent ranges bookend $s^\mathrm{on}=0.5$, the boundary between full guidance and degradation. Because the loss of guidance is the only signal observable through the microscope, subjects seemed to gravitate towards this degradation boundary.

\section{Conclusions and Future Work}

In this work, we present an optimization-based active constraint framework for handheld micromanipulation that provides task guidance and enforces safety simultaneously. Guidance constraints are expressed in the objective with their bandwidth held invariant to hand pose and robot configuration, while safety constraints are enforced with CBFs in a feedforward manner against predicted hand motion. In the case that actuation budget can no longer support both, guidance degrades gracefully to preserve safety. On a medically relevant artificial task, the proposed formulation decreased the safety violation rate by more than two orders of magnitude against the unaided and tremor-canceling baselines, while recovering the 3$\times$ accuracy loss incurred by enforcing safety alone.

In future work, we plan to vary the safety constraint margin online to adapt to different users, fatigue, etc. We also plan to extend evaluation to realistic tasks with richer constraint sets, on operators trained to convergence to remove the learning effects that confound the comparison above.






\bibliographystyle{IEEEtran}
\bibliography{references}

\end{document}